\documentclass[sigconf]{acmart}
\usepackage{enumitem}

\setcopyright{none}
\renewcommand\footnotetextcopyrightpermission[1]{}
\AtBeginDocument{%
  \providecommand\BibTeX{{%
    \normalfont B\kern-0.5em{\scshape i\kern-0.25em b}\kern-0.8em\TeX}}}

\begin{document}

\title{Modelling graph dynamics in fraud detection with ``Attention"}

\author{
    Susie Xi Rao\textsuperscript{\rm 1}, Clémence Lanfranchi\textsuperscript{\rm 1}, Shuai Zhang\textsuperscript{\rm 1}, Zhichao Han\textsuperscript{\rm 2}, Zitao Zhang\textsuperscript{\rm 2}, Wei Min\textsuperscript{\rm 2},  Mo Cheng\textsuperscript{\rm 2}, Yinan Shan\textsuperscript{\rm 2}, Yang Zhao\textsuperscript{\rm 2},  Ce Zhang\textsuperscript{\rm 1}
\\
}

\affiliation{
    \textsuperscript{\rm 1}ETH Zurich
    \hspace{1cm}   
    \textsuperscript{\rm 2}Ebay China \\
    \{raox, clemence.lanfranchi, shuazhang, ce.zhang\}@inf.ethz.ch \\ 
    \{zhihan, zitzhang, wmin, mocheng, yshan, yzhao5\}@ebay.com
}


\renewcommand{\shortauthors}{Rao and Lanfranchi, et al.}
\renewcommand{\authors}{Susie Xi Rao, Clémence Lanfranchi, Shuai Zhang, Zhichao Han, Zitao Zhang, Wei Min,  Mo Cheng, Yinan Shan, Yang Zhao,  Ce Zhang}

\begin{abstract}
At online retail platforms, detecting fraudulent accounts and transactions is crucial to improve customer experience, minimize loss, and avoid unauthorized transactions. Despite the variety of different models for deep learning on graphs, few approaches have been proposed for dealing with graphs that are both heterogeneous and dynamic. In this paper, we propose DyHGN (\textbf{D}ynamic \textbf{H}eterogeneous \textbf{G}raph Neural \textbf{N}etwork) and its variants to capture both temporal and heterogeneous information. We first construct dynamic heterogeneous graphs from registration and transaction data from eBay. Then, we build models with diachronic entity embedding and heterogeneous graph transformer. We also use model explainability techniques to understand the behaviors of DyHGN-* models. Our findings reveal that modelling graph dynamics with heterogeneous inputs need to be conducted with ``attention" depending on the data structure, distribution, and computation cost.
\end{abstract}




\settopmatter{printfolios=true}
\maketitle

\section{Introduction}
Fraud detection is an important task for e-commerce platforms. They are dealing with a large amount of user activities daily, so it is crucial to minimize all sorts of risks involved in online activities, be they account registrations, user interactions, or user transactions. There are two aspects one can model in the aforementioned applications: structural and temporal. Structural refers to the relations between the nodes involved in the activities, temporal the evolution of nodes and edges over time.


We take the scenario of suspicious massive registrations as an illustration. First, abusive accounts are usually interlinked. For example, they may share the same phone number or be registered with the same IP address. They naturally form a graph with heterogeneous types of nodes (e.g., accounts, transactions, IP address, email, phone number).  Second, we want to capture suspicious accounts and transactions within a certain time frame. Indeed, studying the patterns of suspicious registrations shows that, because fraudsters tend to abuse accounts when they are recently registered, the temporal dynamic is a critical factor for detection. 

Take another application scenario we have in e-commerce. We want to understand how the accounts evolve over time in terms of risk scores. Some accounts might be hacked and stolen, some might engage in suspicious transactions after a while, others might be part of a ring attack to the platform. Therefore, it is meaningful to add a temporal aspect in monitoring account behaviors. 

Graph Neural Networks (GNNs) aim to learn a representation vector for each node based on the graph structure. It has been shown in recent studies of fraud detection that GNN-based methods are powerful in flagging frauds (c.f.~\citet{li2019spam, wen2020asa, zhang2019key, liu2018heterogeneous, wang2019fdgars, weber2019anti, ma2018graphrad, liu2019geniepath, liang2019uncovering, rao2021xfraud, lu2021graph, wang2021apan}).

In this paper, we propose DyHGN model and their variants built with diachronic embeddings \cite{goel2020diachronic} and heterogeneous graph transformer (HGT) \cite{hu2020heterogeneous}. DyHGN stands for \textbf{D}ynamic \textbf{H}eterogeneous \textbf{G}raph Neural \textbf{N}etwork. We then apply the DyHGN-* models to three use cases highly relevant in eBay: (a1) detecting suspicious account registration, (a2) flagging risky transactions, and (a3) identifying risky accounts. This allows us to gain insights on modelling dynamic heterogeneous graphs in real-world applications. Hence, we are willing to share those insights with the audiences in both academia and industry. 

\begin{figure*}[!t]
    \centering
    \includegraphics[width=0.95\textwidth]{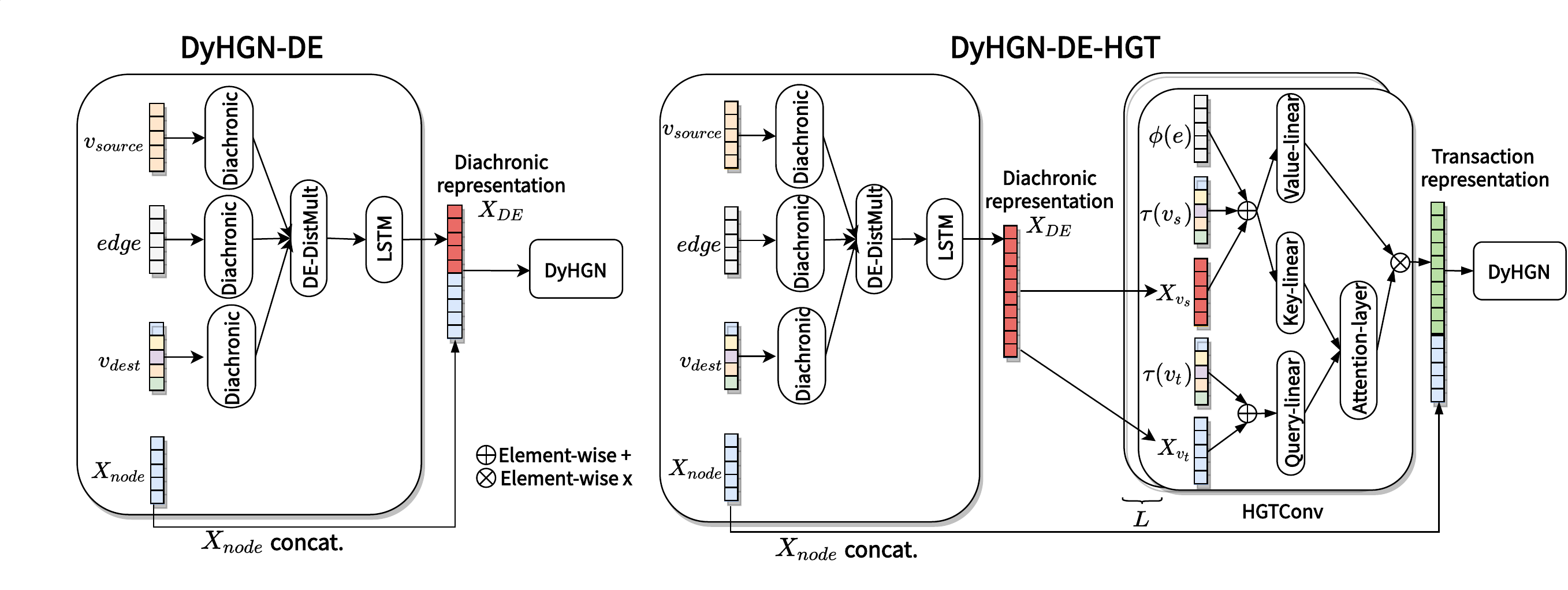}
    \caption{Architecture of DyHGN-DE and DyHGN-DE-HGT.  ``DE": diachronic embedding, ``HGT": heterogeneous graph transformer.}
    \label{fig:DyHGN_variants}
\end{figure*}

First, we discuss DyHGN (Figure~\ref{fig:DyHGN}), which integrates both structural and temporal information by putting them into one graph. We then extend the vanilla DyHGN model with a diachronic entity embedding function with LSTM (DyHGN-DE, Figure~\ref{fig:DyHGN_variants} (left)), which provides the characteristics of entities at any point in time. We also investigate the benefits of capturing the dynamic heterogeneous relation patterns with self-attention based HGT~\cite{hu2020heterogeneous} (DyHGN-DE-HGT, Figure~\ref{fig:DyHGN_variants} (right)). We resonate to the finding in \citet{lv2021we} on the comparison between heterogeneous GNNs (HGNNs) and GAT after having explored the datasets in fraud detection with dynamics. Furthermore, we have discussed the uneven distribution of node labels (uneven proportion of risky labels across time) in the datasets and their influences on the classification performances. Interestingly, we show DyHGN-* models are powerful in learning long-term dependencies from the past, esp. when the label distribution over time fluctuates largely. Last but not least, we use model explainability (Shapley values) in ML models with graph-derived features to gain insights about DyHGN-* models. 

Our key contributions in this work are:

\begin{itemize}[leftmargin=\parindent]
    \item We have developed a prototype (DyHGN) that models the account dynamics in a heterogeneous transaction graph. On top of DyHGN, we built other variants with diachronic entity embeddings \cite{goel2020diachronic} and HGT \cite{hu2020heterogeneous} modules and evaluate their performances on real-world industrial datasets.
    \item We benchmark multiple GNNs against three datasets from an e-commerce retail platform. 
    \item We use ML models trained on graph-derived features and use the insights to better explain DyHGN-* models.
    \item We open-source the codebase and data\footnote{Note that \textit{eBay-small} dataset in \cite{rao2021xfraud} (desensitized transaction records, denoted as xFraud here) is used in this work. Please contact the authors for DATA USE AND RESEARCH AGREEMENT (eBay) and obtain the usage rights of eBay-small dataset. In the long run, it would be possible to share the MassReg dataset after the legal review at eBay.} to facilitate further research in these directions. The codebase/data are made available on this GitHub repository: \url{https://github.com/DS3Lab/DyHGN}.
\end{itemize}


\section{Research Question and Methodology}
In this section, we first present our research question, the problem definition (Sec.~\ref{sec:problem-def}) of fraudulent accounts and transactions, from the perspective
of modeling a dynamic heterogeneous graph. Then, we explain the graph construction (Sec.~\ref{sec:graph-construction}), our DyHGN architecture (Sec.~\ref{sec:DyHGN}), the diachronic embedding component (Sec.~\ref{sec:diachronic-emb}), and the DyHGN-* variants (Sec.~\ref{sec:DyHGN-variant}).

\subsection{Research Question}
In relation to daily use cases at eBay, we aim to address the following question: how can we develop an end-to-end framework to leverage the entities and time information available in the following scenarios (a1) detect suspicious account registration, (a2) flag risky transactions, and (a3) identify risky accounts?

\subsection{Problem Definition}
\label{sec:problem-def}

As listed above, we study three application scenarios in the paper which would benefit from leveraging both dynamic and heterogeneous information from transaction graphs: (a1) detect suspicious account registration, (a2) flag risky transactions, and (a3) identify risky accounts. Here, we take scenario (a1) as an example to illustrate how we build a dynamic heterogeneous graph and a model around the use case. The other two use cases (a2 and a3) follow a similar design. 

Massive account registration refers to a process where a user or an organization can add a group of accounts simultaneously\footnote{See this illustration for an experience of massive user registration, https://hellohelp.gointegro.com/en/articles/1488462-massive-users-registration (last accessed: Nov. 9, 2020).}. This functionality is enabled on many e-commerce websites to speed up batch registration for a group of accounts. By providing the entities required in a registration template, a user can create within a short period several accounts. Apart from template-based operations, a bot can also be built to submit forms multiple times with generic information. In practice, a fraudster can make use of this functionality on an e-commerce website to register a set of accounts for multiple purposes: using stolen financial means, providing fake reviews, redeeming coupons and purchasing using an abusive gift card, etc. Oftentimes, these registrations share some common patterns of disguise.  

Think of the critical entities involved in massive account registration. A credit card might be stolen and used by one of the newly registered accounts. A suspicious account can be registered with a common shipping address such as a warehouse, which is a disguise of suspicious activity. These accounts are oftentimes registered using email addresses with spam patterns, also using telephone numbers that are listed as spam calls by third-party collaborators for risk detection. We can also detect a chain of illegal activities within an e-commerce platform: an account has been hacked and a fraudster registers a batch of new accounts to use the gift cards and financial instruments that are bundled with the hacked account. 

Furthermore, the time dimension is crucial in massive registration. Based on the manual analysis in the business unit on suspicious massive registrations, usually within three months after registration, fraudster activities occur. An effective system should be able to link suspicious entities used in the past or uncover suspicious patterns discovered in the past graph snapshots. Later, when such a system is deployed in production, it is expected to provide both feature level detection and graph level detection. Feature level detection is achievable via a rule-based system, while graph level detection is better modeled by an end-to-end model that learns the time-variant representations quickly and effectively.

Formally, a heterogeneous network is defined as a graph $\mathcal{G} = (\mathcal{V}, \mathcal{E}, \mathcal{A}, \mathcal{R})$, where node type mapping function is $\tau : \mathcal{V} \rightarrow \mathcal{A}$ and link type mapping function is $\phi : \mathcal{E} \rightarrow \mathcal{R}$. Each node $v \in \mathcal{V}$ has only one type $\tau(v) \in \mathcal{A}$, and each edge $e \in \mathcal{E}$ has only one type $\phi(e) \in \mathcal{R}$. Each node or edge can be associated with attributes, denoted by $X_{\tau({v})}$. In addition, each node or edge can be labeled, denoted by $y \in \mathcal{C}$. For each time $t$, an entity in $\mathcal{V}$, or a link in $\mathcal{E}$, can be removed from or added into the graph snapshot $G_t$.

\begin{figure*}[ht]
    \centering
    \includegraphics[width=\textwidth]{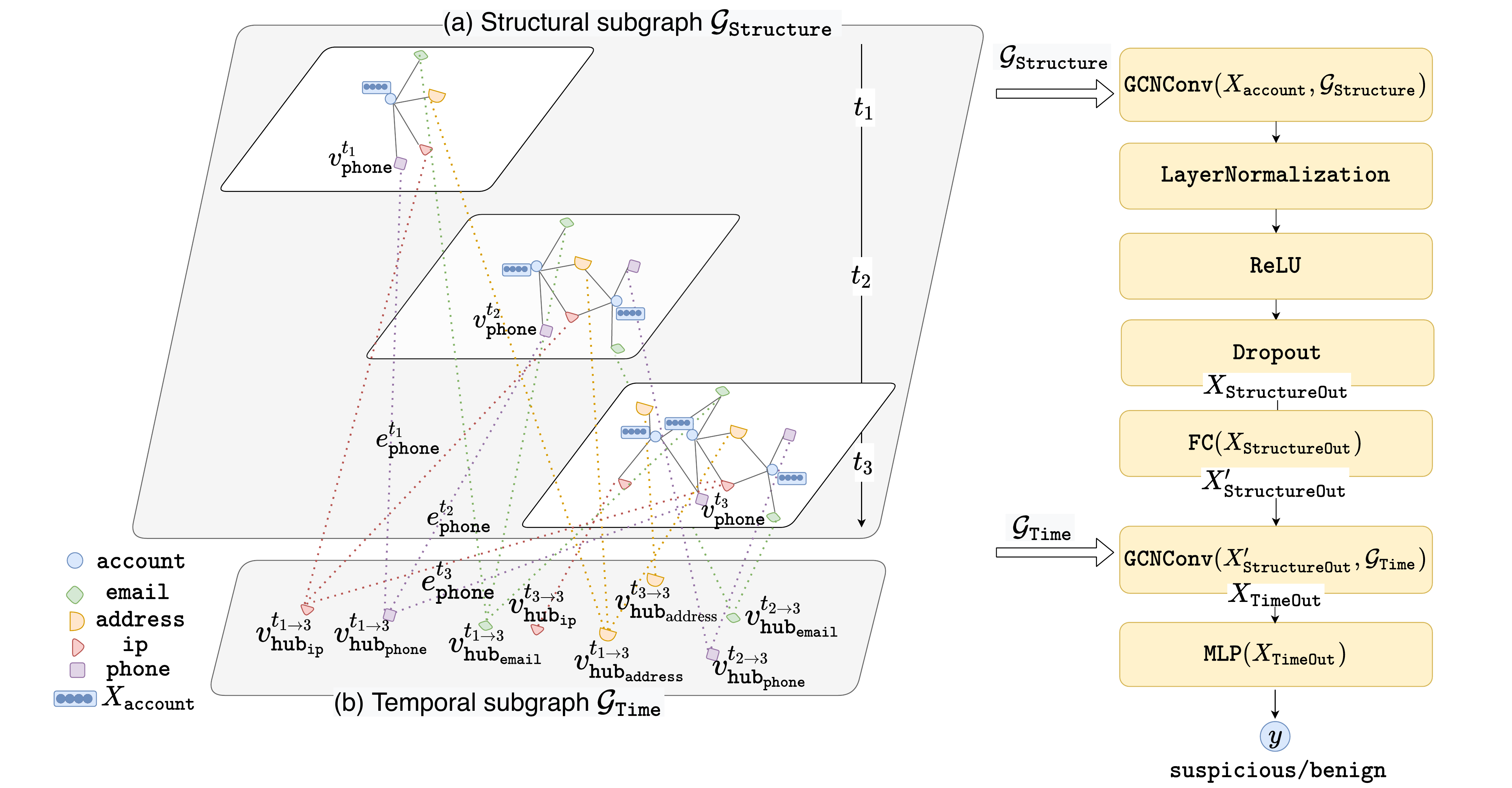}
    \caption{DyHGN: \underline{Dy}namic \underline{H}eterogeneous \underline{G}raph {N}eural \underline{N}etwork for Suspicious Massive Registration.}
    \label{fig:DyHGN}
\end{figure*}

\subsection{Graph Construction} \label{sec:graph-construction}
In our work, we populate the graph with three different applications that we introduce with further details in Sec.~\ref{sec:data}. Again, we take the application of detecting suspicious registrations (a1) as an illustration. 
We formulate it as a binary classification problem in a transductive setting in a heterogeneous graph. Thus, we specify the problem definition from the heterogeneous and temporal aspects as follows. 

\textbf{Heterogeneous.} In a heterogeneous transaction graph $\mathcal{G}$, $v \in \mathcal{V}$ has a type $\tau(v) \in \mathcal{A}$, where $\mathcal{A}:=$ \{\textit{account}, \textit{address}, \textit{IP}, \textit{phone}, \textit{email}\}, referring to account, registration address, IP address, phone, email, respectively\footnote{For this study, we choose these attributes because they reflect typical patterns of massive registration activities as reported by our business unit. Other types of attributes such as device type are incorporated in the in-house risk detection system.}. If an account uses a linking entity in $\{$\textit{address, IP, phone, email}$\}$, we put an edge between the account node and this linking entity in the heterogeneous graph. Each $account$ node carries node attributes provided by a risk identification system. Each account ID is flagged as benign or suspicious. We use these flags as labels in our binary classification.

\textbf{Temporal.} In general, there are two ways (\textit{\textbf{static}} and \textit{\textbf{dynamic}}) to incorporate the time dimension into a heterogeneous graph using accounts and their linking entities. 

\textbf{\textit{Static}.}
One can either construct different static snapshots of a graph with all involving entities for each time $t$. This means a detection system has to process all snapshots $\{G_1,\dots,G_T\}$, where $T$ is the number of time steps. In each snapshot, a graph is constructed. This is the graph construction DySAT~\cite{sankar2020dysat} employs. However, as analyzed in Section~\ref{sec:lit}, DySAT is not suitable for our application because not all entities are appearing in every snapshot. If we break down the graph into snapshots, we will lose many time-dependent linkages and because many accounts have not existed and hence cannot be linked to any entity. Another reason is that newly registered accounts that are suspicious will only be abused within the $T$ weeks ($T$ is usually small, say 1-12). Consequently, we only need to encode the linking entities such as IP addresses and registration addresses for those periods. Hereafter, we propose an alternative of constructing the graph for our applications and present it below.

\textbf{\textit{Dynamic.}}
Unlike treating a heterogeneous graph as a set of static snapshots, we unroll the time snapshots into one graph to incorporate nodes and edges that (dis)appear overtime. TIMESAGE~\cite{shekhar2020entity} discusses is a similar design: a time-dependent network representation. We build a graph that tracks entities which could be present or absent in all snapshots. Entities such as phone numbers and addresses are present in all snapshots, but accounts can only exist after certain snapshots, i.e., after the accounts have been created.

We have two main components in our dynamic heterogeneous transaction network, a structural subgraph (Figure~\ref{fig:DyHGN} (a)), and a temporal subgraph (Figure~\ref{fig:DyHGN} (b)). Here we discuss the design of these two network components and the intuitions behind the design. The structural subgraph is designed to reflect the linkages among various types of entities. The nodes are accounts and attributes used in account registration. For instance, if two accounts are registered using the same IP, an edge is added from account 1 to this IP, and another edge is added from account 2 to this IP as well. Consequently, account 1 and account 2 are linked via this IP in this way (see Figure~\ref{fig:DyHGN} (a)). Hence, the structural subgraph captures the relationships among the entities and allows us to uncover the patterns in account registration. Now let us talk about the temporal subgraph that builds on top of the structural components in each time $t$. For each time $t$, we observe a structural subgraph constructed as we describe above. Then, we add temporal edges from the structural nodes to a node called $v_{hub}$. These structural nodes in different timestamps represent the identical entities in each time $t$. We also index $v_{hub}$ with the timestamp(s) when it appears. In the example shown in Figure~\ref{fig:DyHGN}, the nodes $v_{hub_{phone}}^{t_{1}}$, $v_{hub_{phone}}^{t_{2}}$, $v_{hub_{phone}}^{t_{3}}$ are connected to a node $v_{hub_{phone}}^{t_{1\rightarrow3}}$ via the temporal edges $e_{phone}^{t_{1}}$, $e_{phone}^{t_{2}}$, and $e_{phone}^{t_{3}}$. From the $v^t$ nodes to the $v_{hub}$ node we have a star graph that represents if an entity has appeared in time $t$ or not. We denote the unrolled dynamic heterogeneous graph as $\mathcal{G}_T$, where all the edges and nodes appearing from $\{1, \dots, T\}$ are present. $\mathcal{G}_T$ is composed of $\mathcal{G}_{Structure}$ and $\mathcal{G}_{Time}$.

\begin{table}[!t]
    \centering
    \resizebox{\linewidth}{!}{\begin{tabular}{l|l}
    \toprule
         \textbf{Notation} & \textbf{Description}  \\
    \midrule
         $G_t$ & a graph snapshot at time $t$ \\
         \midrule
         $T$ & the number of time steps \\
         \midrule
         $\mathcal{G}_T$ & a dynamic heterogeneous graph from timestamps $\{1, \dots, T\}$ \\
         \midrule
         $\mathcal{G}_{Structure}$ & structural subgraph in Figure~\ref{fig:DyHGN} (a) \\
         \midrule
         $\mathcal{G}_{Time}$ & temporal subgraph in Figure~\ref{fig:DyHGN} (b) \\
         \midrule
         $X_{account}$ & account level features \\
         \midrule
         $X_{StructureOut}$ & the output of structural message passing \\
         \midrule
         $X^{\prime{}}_{StructureOut}$ & the output of FC transformation of $X_{StructureOut}$ \\
         \midrule
         $X_{TimeOut}$ & the output of structural message passing \\
         \midrule
         $X_{DE}$ & the diachronic entity embeddings \\
         \midrule
         $e_i^t$ & an edge in each time $t$ for type $i$, e.g., $e_{phone}^{t_1}$ \\
         \midrule
         $v_i^t$ & a node in each time $t$ for type $i$, e.g., $v_{phone}^{t_1}$ \\
         \midrule
         $v_{{hub}_i}^{{t_{i\rightarrow j}}}$ & a hub node in from $t_i$ to $t_j$ for type $i$, e.g., $v_{hub_{phone}}^{t_{1 \rightarrow 3}}$ \\
         
    \bottomrule
    \end{tabular}}
    \caption{Notations.}
    \label{tab:notations}
\end{table}

\subsection{DyHGN Architecture}
\label{sec:DyHGN}
Now we present the architecture of DyHGN, which is applicable in all three scenarios discussed in this paper. DyHGN is composed of two subgraphs, a structural subgraph $\mathcal{G}_{Structure}$ to capture the relations between different types of entities and a temporal subgraph $\mathcal{G}_{Time}$ to capture the dynamic aspect of the entities and to determine whether an entity appears in time $t$ or not. Temporal edges are added from the structural nodes to their timestamp-indexed counterparts in the temporal subgraph. For each subgraph, a graph convolutional network (GCN) layer is used to learn the message passing between the nodes.

\textbf{Structural message passing.}
The first GCN layer takes as input the structural subgraph $\mathcal{G}_{Structure}$ and the node level features $X$ (e.g., $X_{account}$ in MassReg). Only the nodes to classify have features, while the other node types do not have initial feature values and are initialized with zero vectors. Then, we use a feedforward layer to connect the message passing between the structural and temporal parts. The output is layer-normalized and fed into a nonlinear transformation using ReLU as the activation function. Finally, a Dropout layer is applied to regularize and avoid overfitting. The output of this step is $X^{\prime{}}_{StructureOut}$.

\textbf{Temporal message passing.}
The second GCN layer takes as input the temporal subgraph $\mathcal{G}_{Time}$ and $X_{StructureOut}$. More specifically, in order to obtain $\mathcal{G}_{Time}$, we create temporal nodes where each linking entity $v$ of type $i$ appearing in time snapshot $t$ gives us a temporal node $v_i^t$. Similarly, we add temporal edges that connect the original node from the structural subgraph to its temporal counterparts in each time stamp. After the GCN layer, we apply dropout, layer normalization, and ReLU.

\textbf{Prediction.}
There can be multiple convolution layers of structural and temporal message passing (see Table~\ref{tab:hyperparam} about \text{n\_layers}). The output from the final convolution is fed into a feedforward connected (FC) network. We then apply layer normalization, ReLU, and dropout before calculating a risk score (aka probability) for the input node, using softmax. Based on the risk score, a label is generated (this part is summarized under MLP in Figure~\ref{fig:DyHGN}). For the xFraud datasets, the loss function is a cross entropy of the true label, whereas for the MassReg dataset, the loss function is an average of the binary cross entropy and the multi-class cross entropy, because accounts are also labeled with different types of risk levels (see MassReg dataset in Sec.~\ref{sec:data}).

\subsection{DyHGN-* Variants}\label{sec:DyHGN-variant}

It is worth noting that while DyHGN is applied to heterogeneous graphs, its convolution layer uses GCN, a convolution designed for homogeneous graphs. We further explore methods to modify the DyHGN model with better representations of temporal and structure modules. We introduce two model variants, DyHGN-DE and DyHGN-DE-HGT, which modify the temporal and structural aspects, respectively. In the subsequent sections, we first introduce the diachronic embedding module, which enables us to model the temporal aspects of an entity.


\subsubsection{\textbf{Diachronic Embedding}}
\label{sec:diachronic-emb}
To take the temporal dynamic further into account, we equip DyHGN model with \textbf{d}iachronic entity \textbf{e}mbedding following \citet{goel2020diachronic}, which gives us DyHGN-DE. This approach builds upon  static entity embeddings and proposes an alternative which takes time as input as well, to provide the characteristics of entities at any point in time. The use of diachronic embedding has proved beneficial in temporal knowledge graph (KG) completion.  In our work, we construct a diachronic embedding based on the DistMult~\cite{yang2014embedding} score function, which we denote as DE-DistMult and provide its formal definitions below.
Let $\mathcal{V}$ be a finite set of entities, $\mathcal{R}$ be a set of relation types, and $\mathcal{T}$ be a finite set of timestamps.

\begin{definition}[Diachronic entity embedding]
\hspace{2cm}\\
A diachronic entity embedding, denoted as $\texttt{DEEMB}$, is a function which maps every pair $(v,t)$, where $v\in \mathcal{V}$ and $t\in\mathcal{T}$, to a hidden representation. 
\end{definition}

\begin{definition}[DE-DistMult]\hspace{4cm}\\
In the DE-DistMult scoring function $\phi$, 
\begin{itemize}[leftmargin=\parindent]
    \item we define for the nodes their entity embedding $\texttt{DEEMB}(v, t) = (\mathbf{z}^t_v)$ for every $v\in\mathcal{V}$ where $\mathbf{z}^t_v \in \mathbb{R}^d$,
    \item we define for the edges their relation embedding $\texttt{REMB}(r)=(\mathbf{z}_r)$ for every $r\in\mathcal{R}$ where $\mathbf{z}_r \in \mathbb{R}^d$, and $\phi(v,r,u)=\langle\mathbf{z}^t_v,\mathbf{z}_r,\mathbf{z}^t_u\rangle$. 
\end{itemize} 
\end{definition}

Finally, we define the diachronic embedding $\mathbf{z}^t_v$
\begin{equation}\label{eq:1}
    \begin{array}{l}
        \mathbf{z}^t_v[n] = \begin{cases} 
            \mathbf{a}_v[n]\sigma(\mathbf{w}_v[n]t+\mathbf{b}_v[n]), & \text{if $1\leq n \leq \gamma d$ }\\
            \mathbf{a}_v[n], & \text{if $\gamma d < n \leq  d$}
		 \end{cases}
    \end{array}
\end{equation}

where $\mathbf{z}^t_v[n]$ represents the $n$\textsuperscript{th} element of vector $\mathbf{z}^t_v$, $\mathbf{a}_v\in \mathbb{R}^d$ and $\mathbf{w}_v, \mathbf{b}_v \in \mathbb{R}^{\gamma d}$ are (entity-specific) vectors with learnable parameters $\gamma$ and $\sigma$ is an activation function. $0 \leq \gamma \leq 1$ is a hyperparameter controlling the percentage of temporal features. The first $\gamma d$ elements of the vector in Equation (\ref{eq:1}) capture temporal features and the other $(1-\gamma)d$ elements capture static features. We use \textit{sine} as the activation function. 

\textbf{Diachronic embedding vs.~temporal subgraph.} While DyHGN constructs a temporal subgraph for each time $t$, the diachronic embedding takes an entity and its timestamp as input and provides a hidden representation for the entity at that time, where the parameters of hidden representations are learned from the data. We obtain a diachronic representation of each node that has temporal and structural information, by computing the DE-DistMult scores for all relations in which an entity is involved and average them.

Now, we explain in detail how we incorporate the diachronic embedding in a heterogeneous setting into our DyHGN model. We denote the node level features as $X$.
For every edge $(v,r,u)$ in the structural graph with $v\in\mathcal{V}$, $u\in\mathcal{V}$ and $r\in\mathcal{R}$, we compute the DE-DistMult score $\phi(v,r,u)=\langle\mathbf{z}^t_v,\mathbf{z}_r,\mathbf{z}^t_u\rangle $ of their diachronic embedding. Then, we use an LSTM layer (Long Short Term-Memory) to aggregate the scores of the edges that are linked to a same node. Finally, we concatenate the obtained diachronic embedding of each node to the $X$ table. The output after this operation is denoted as $X_{DE}$. Using an LSTM deals nicely with the variable number of edges that a node can be involved in. 

\subsubsection{\textbf{DyHGN-DE and DyHGN-DE-HGT}}
Now, we explain two DyHGN-* variants. (1) (\textbf{DyHGN-DE}) We directly apply the DyHGN model to our new input $X_{DE}$, $\mathcal{G}_{Structure}$, and $\mathcal{G}_{Time}$. (2) (\textbf{DyHGN-DE-HGT}) We first apply a Heterogeneous Graph Transformer (HGT) layer \cite{hu2020heterogeneous} to $\mathcal{G}_{Structure}$. The input to the HGT layer is $X_{DE}$ and $\mathcal{G}_{Structure}$, the output $X_{HGT}$. HGT has node- and edge-type dependent parameters, which makes our graph convolution layer also heterogeneous. Finally, we apply the DyHGN model with inputs $X_{HGT}$, $\mathcal{G}_{Structure}$, and $\mathcal{G}_{Time}$.

\begin{table}[!t]
    \centering
    \resizebox{0.7\linewidth}{!}
    {\begin{tabular}{c|r|c|r}
        \toprule
        \textbf{Node Type} & \textbf{Count} & \textbf{Edge Type} & \textbf{Count} \\
        \midrule
         Account ID & 111,691 & Temporal Edge & 135,614\\
         Address & 7,221 & Account ID - Address & 29,217 \\
         IP Address & 6,762 & Account ID - IP Address & 104,719\\
         Phone & 4,958 & Account ID - Phone & 18,542 \\
         Email & 134 & Account ID - Email & 608 \\
         \midrule
         TOTAL & 130,766 & TOTAL & 288,700 \\
        \bottomrule
    \end{tabular}}
    \caption{Node \& Edge Count/Types for the MassReg Graph.}
    \label{tab:MassReg}
\end{table}

\begin{table}[!t]
    \centering
    \resizebox{0.8\linewidth}{!}
    {\begin{tabular}{c|r|c|r}
        \toprule
        \textbf{Node Type} & \textbf{Count} & \textbf{Edge Type} & \textbf{Count} \\
        \midrule
        \multicolumn{4}{c}{\textbf{xFraudTxn}}\\
        \midrule
         Transaction & 207,749 & Temporal Edge & 350,846 \\
         Account ID & 28,815 & Transaction - Account ID & 124,818\\
         Payment token & 22,273 & Transaction - Payment & 102,569 \\
         Address & 7,138 & Transaction - Address & 199,957 \\
         Email & 25,878 & Transaction - Email & 185,560 \\
         \midrule
         TOTAL & 291,853 & TOTAL & 963,750 \\
         \midrule
        \multicolumn{4}{c}{\textbf{xFraudAccount}}\\
        \midrule
         Account ID & 28,815 & Temporal Edge & 488,654 \\
         Transaction & 207,749 & Account ID - Transaction & 124,818 \\
         Payment token & 22,273 & Account ID - Payment token & 101,728  \\
         Address & 7,138 & Account ID - Address & 117,158 \\
         Email & 25,878 & Account ID - Email & 124,796 \\
        \midrule
         TOTAL & 291,853 & TOTAL & 957,154 \\
        \bottomrule
    \end{tabular}}
    \caption{Node \& Edge Count/Types  for the xFraud Graphs.}
    \label{tab:xFraud}
\end{table}
\section{Experiments, Evaluation, and Discussions}
In this section, we introduce our datasets and their preprocessing (Sec.~\ref{sec:data}), show the experimental setups (Sec.~\ref{sec:baseline} and~\ref{sec:implementation}), evaluate the model performances (Sec.~\ref{sec:eval}) with ablation studies (Sec.~\ref{sec:gnn-vs-ml}), and discuss our findings (Sec.~\ref{sec:discussion}).

\subsection{Dataset, Preprocessing, and Evaluation Metric}
\label{sec:data}

We use two eBay data sources (MassReg and xFraud\footnote{The \textit{ebay-small} dataset in \cite{rao2021xfraud}.}), from which we derive three application scenarios: (a1) detecting suspicious account registration (MassReg), (a2) flagging risky transactions (xFraudTxn), and (a3) identifying risky accounts (xFraudAccount). In Tables \ref{tab:MassReg} and \ref{tab:xFraud}, we report node and edge types and their counts.

\textbf{MassReg dataset.} This dataset was sampled from the real-time account registration logs from September to December in 2019. We create a heterogeneous graph where each node $v\in\mathcal{V}$ has a type $\tau(v)\in \mathcal{A}$ where $\mathcal{A} := \{account, adress, IP, phone, email\} $, referring to account, registration address, IP address, phone, email, respectively. Account features are 264-dimensional vectors, encoding risk features generated by an in-house risk detection system. Each account is flagged as benign or suspicious. The labels are either automatically generated by rule-based filtering on transaction behaviors (e.g., riskiness, payment rejection, chargeback, abusive buyers, etc.) or by manual annotations deducing from the registration rules. The dataset is balanced with about 50\% of suspicious registrations, yet these registrations are not even across time, with a peak of suspicious registrations  during the first two weeks of our time window (c.f. Figure \ref{fig:risky} in Appendix \ref{app:dist}). For a more detailed description of MassReg, refer to \cite{rao2020suspicious}.

\textbf{xFraud dataset.}
This dataset contains financial transactions carried on the platform across 6 weeks. Transaction records have a rich set of relations, which enables us to derive two application scenarios: (a2) detecting risky transactions and (a3) flagging suspicious accounts. For this, we create two heterogeneous graphs where, for both graphs, each node $v\in\mathcal{V}$ has a type $\tau(v)\in \mathcal{A}$ where $\mathcal{A} := \{txn, pmt, email, addr, buyer\} $, referring to transaction, payment token, email, shipping address, buyer, respectively. For a more detailed description of xFraud, refer to \citet{rao2021xfraud} (the ebay-small dataset in \cite{rao2021xfraud}).  We present two graphs as follows.
\begin{itemize}[leftmargin=\parindent]
    \item \textbf{xFraudTxn}. If a transaction has a relation with another type of node in $\{pmt, email, addr, buyer\}$, we put an edge between two nodes. Each $txn$ carries node attributes as a 114-dimensional vector, encoding different features, e.g., item type, device type, and IP, from which the transaction was made. Each transaction is flagged legit or fraud, which we want to predict.
    \item \textbf{xFraudAccount}. If an account has a relation with another type of node in $\{txn, pmt, email, addr\}$, we put an edge between two nodes. Each $buyer$ has features which are initialized as an average of the transactions' attributes to which they are connected.
\end{itemize}
Different from the unevenly distributed labels across time in the MassReg graph, we see an even distribution of labels in the xFraud dynamic graphs. We provide the detailed graphs of labels distribution across time in Figure \ref{fig:risky} (Appendix \ref{app:dist}). To prevent data leakage, accounts' labels are not used in the transaction classification task, and reciprocally for transactions' labels in the account classification. Since we only have transactions timestamps in this dataset, we define the date of account creation as the day of their first (possible) transaction. These datasets are highly imbalanced, with 1.5\% of fraudulent transactions and 3.5\% of fraudulent accounts (note that we report the average percentage across time stamps).

For all three application scenarios, we perform a binary classification task on nodes in a transductive setting (similar to \citet{rossi2020temporal}), where all edges are available during training and node labels are split chronologically in the ratio of 70\%-10\%-20\%.\footnote{While the test set was split chronologically, the train/validation split was performed randomly, which can be questionable since we are mainly interested in the temporal aspect of the graph. For this reason, we also evaluate the DyHGN model with a chronological Train/Val split. We observed a slight decrease in performance for all models, but it did not change the score significantly.} Since we have binary labels, we use a popular evaluation metric in risk management: the average precision (AP), which corresponds to the area under the precision-recall curve. We prefer to choose this metric over the Area Under the ROC Curve (ROC-AUC) metric since we care more about correctly classifying the positive class, and we also deal with imbalanced data. However, we still report the AUC scores in Appendix~\ref{app:auc} for the interested reader.

\subsection{Baselines}
\label{sec:baseline}
We select our baseline models to be simple GNN models such as graph convolutional network (GCN) and graph attention network (GAT). They are designed to model homogeneous graphs, but their performances on heterogeneous graphs has been redeemed by \citet{lv2021we}, who show that vanilla GAT can even outperform existing HGNNs in most cases. We also compare our models to Simple-HGN~\cite{lv2021we}.
We report below the detailed settings of each model.
\begin{itemize}[leftmargin=\parindent]
    \item GCN is one of the first GNN models and uses average aggregation from neighbors with the objective to learn a shared "convolution kernel" which could be applied to every part of the graph in order to absorb the information from neighbors to the node.
    \item GAT uses an attention mechanism to perform a weighted aggregation from one-hop neighbors. Along with GCN, these homogeneous GNNs can handle heterogeneous graphs by simply ignoring node and edge types.
    \item Simple-HGN. Starting from the GAT model, it includes edge type information into attention calculation through a learnable edge-type embedding, making it possible to model heterogeneous graphs. The model is also enhanced with residual connection and $L_2$ normalization on the output embedding.
\end{itemize}

\subsection{Implementation Details}
\label{sec:implementation}

All models are implemented in \textit{Pytorch} and were run on NVIDIA TITAN X  with 12G of memory. Concerning the diachronic embedding, since the timestamps in our MassReg dataset are dates rather than single numbers, we apply the temporal part of Equation (1) to week and day separately (with different parameters) and thus obtain two temporal vectors. Then we take an element-wise sum of the resulting vectors, to obtain a single temporal vector. Intuitively, this can be viewed as converting a date into a timestamp in the embedded space. We report in Table \ref{tab:hyperparam} (Appendix \ref{app:hyper}) the hyperparameters chosen for our models after hyperparameter tuning on the validation set. We train our models on 5 different seeds to get the average scores.

\subsection{Performance Evaluation}
\label{sec:eval}

Now we present the results of DyHGN-* models on three datasets. 
\subsubsection{\textbf{Overall Performances: AP scores}}
We report the AP scores of our models and their baselines in Table \ref{tab:result}. Comparing the results across models, we make observations as follows:
\begin{enumerate}[leftmargin=\parindent]

\item \textbf{The power of GAT.} GAT is a very strong model (which is in line with the findings in \citet{lv2021we}), esp. in the evenly distributed datasets across time like xFraudTxn and xFraudAccount. 


\item \textbf{The power of temporal modeling.} We use ``DE" module to model time-related entity embeddings across different timestamps. In MassReg (a dataset with an uneven distribution), our DyHGN and DyHGN-DE models have stronger performances over GAT. This is because the message passing from the previous time stamps is better learned in DyHGN-(DE) models with a temporal focus (node/edge appearance and disappearance). Also, the LSTM layer helps learning long-term dependencies from the past. In MassReg where the temporal patterns fluctuate largely across time, DyHGN-DE is very suitable to model the dynamics. On xFraud datasets where the temporal patterns do not vary largely overtime, DyHGN-DE does not have strong performance. Another reason that DyHGN-DE is doing better on MassReg might come from the fact that we were able to have a diachronic embedding of larger size than for the other datasets (due to GPU memory limit), shown in Table \ref{tab:hyperparam} (Appendix \ref{app:hyper}). Therefore, using larger diachronic embedding size could be beneficial.


\item \textbf{The power of heterogeneous graph transformer.}
In DyHGN-DE-HGT, we adopt the ``HGT" module for the heterogeneous graph transformer between different types of nodes and edges. We notice that a complex convolution like HGT does not always outperform Simple-HGN~\cite{lv2021we}. 

\item \textbf{Varying performances among DyHGN-* model.} It is worth noting that DyHGN-DE-HGT does not outperform the other two DyHGN-* variants and GAT on all three datasets, this means that modeling the temporal aspects have contributed largely to the prediction. Using a more complex convolution (HGT vs. GCN/GAT) might deteriorate the performances. 
    
\end{enumerate}

\begin{table}[!t]
\centering
\resizebox{\linewidth}{!}{\begin{tabular}{cccc}
\toprule
& MassReg & xFraudTxn & xFraudAccount \\
\midrule

GCN      & 0.7995 $\pm$ 0.0058 & 0.1380 $\pm$ 0.0126 & 0.0606 $\pm$ 0.0039 \\
GAT      & 0.8193 $\pm$ 0.0049 & \textbf{0.1504} $\pm$ 0.0087 & 0.1638 $\pm$ 0.0041 \\
Simple-HGN  & 0.7566 $\pm$ 0.0253 & 0.0615 $\pm$ 0.0069 & \textbf{0.2588} $\pm$ 0.0267 \\
DyHGN   & 0.8239 $\pm$ 0.0086 & 0.0950 $\pm$ 0.0018 & 0.1195 $\pm$ 0.0525 \\
DyHGN-DE  & \textbf{0.8298} $\pm$ 0.0032 & 0.0323 $\pm$ 0.0010 & 0.0625 $\pm$ 0.0038\\
DyHGN-DE-HGT  & 0.8047 $\pm$ 0.0159 & 0.0321 $\pm$ 0.0037 & 0.0669 $\pm$ 0.0095 \\
\bottomrule
\end{tabular}}
\caption{Experiment Results with the Average Precision (AP) score. Our models are denoted as DyHGN-*. ``DE": diachronic embedding,  ``HGT": heterogeneous graph transformer.}

\label{tab:result}
\end{table}

\subsubsection{\textbf{Ablation Studies}}
\label{sec:ablation}
We describe ablation studies on the MassReg dataset (where DyHGN-* outperforms baselines) with several variants of proposed models to provide better understanding of their performances.

\textbf{GCN vs. DyHGN-*.} We compare the two model variants, DE + HGT and DyHGN-DE-HGT. The main difference is that the former has only the structural subgraph using HGT (no temporal subgraph) and the latter one has both subgraphs. The former performs worse than the latter. So the comparison indicates that diachronic embeddings, temporal, and structural subgraphs are all important in modelling graph dynamics. And diachronic embeddings and temporal subgraph capture different aspects of dynamics.

\textbf{Aggregation of diachronic embedding.} So far, we used an LSTM layer in order to aggregate the DE-DistMult scores of the edges linked to a  node. We also tried taking the simple average of all scores, but observed a worsened performance from that variant. This suggests that the model really benefits from the LSTM layer being able to capture long-term dependencies.

\textbf{Adding diachronic embedding for relations.} As already highlighted in \citet{goel2020diachronic}, we hypothesize that relation evolution is negligible (compared with node evolution), therefore, modeling relations with a static, rather than a diachronic embedding suffices. We tested this hypothesis by running our models where relation embeddings are also a function of time. We observed no significant improvement, meaning that the evolution of relations is not helpful in our setting. 
\begin{table*}[!t]
\centering
\resizebox{0.85\linewidth}{!}{
\begin{tabular}{lcccccccccc}
\toprule
          & day         & week       & relations\_address & relations\_ip & relations\_phone & relations\_email & snapshots\_address & snapshots\_ip & snapshots\_phone & snapshots\_email \\
\midrule
0          & 27          & 3          & 0                  & 41            & 0                & 0                & 0                  & 13            & 0                & 0                \\
\textbf{1} & \textbf{63} & \textbf{9} & \textbf{4}         & \textbf{2}    & \textbf{9}       & \textbf{0}       & \textbf{2}         & \textbf{2}    & \textbf{2}       & \textbf{0}       \\
2          & 23          & 3          & 2                  & 28            & 0                & 0                & 2                  & 1             & 0                & 0                \\
3          & 58          & 8          & 0                  & 8             & 0                & 0                & 0                  & 3             & 0                & 0                \\
4          & 20          & 2          & 0                  & 121           & 0                & 0                & 0                  & 10            & 0                & 0    \\
\bottomrule
\end{tabular}}
\caption{Example of Graph-derived Features.}
\label{tab:features}
\end{table*}

\begin{table}[!t]
\centering
\resizebox{0.75\linewidth}{!}{\begin{tabular}{ccc}
\toprule
                               & Global & Incremental \\ \midrule
Random train/test split        & 0.8246          &      0.7953          \\
Chronological train/test split & 0.5598          &      0.5301          \\
\bottomrule
\end{tabular}}
\caption{AP Scores of XGBoost. ``Global": using features computed from all time stamps, ``Incremental": using only features until the time $t$.}
\label{tab:graph_features}
\end{table}

\subsection{Deep Dive into DyHGN-* using ML Models with Graph-derived Features}
\label{sec:gnn-vs-ml}

\begin{figure}[!t]
    \includegraphics[width=0.45\textwidth]{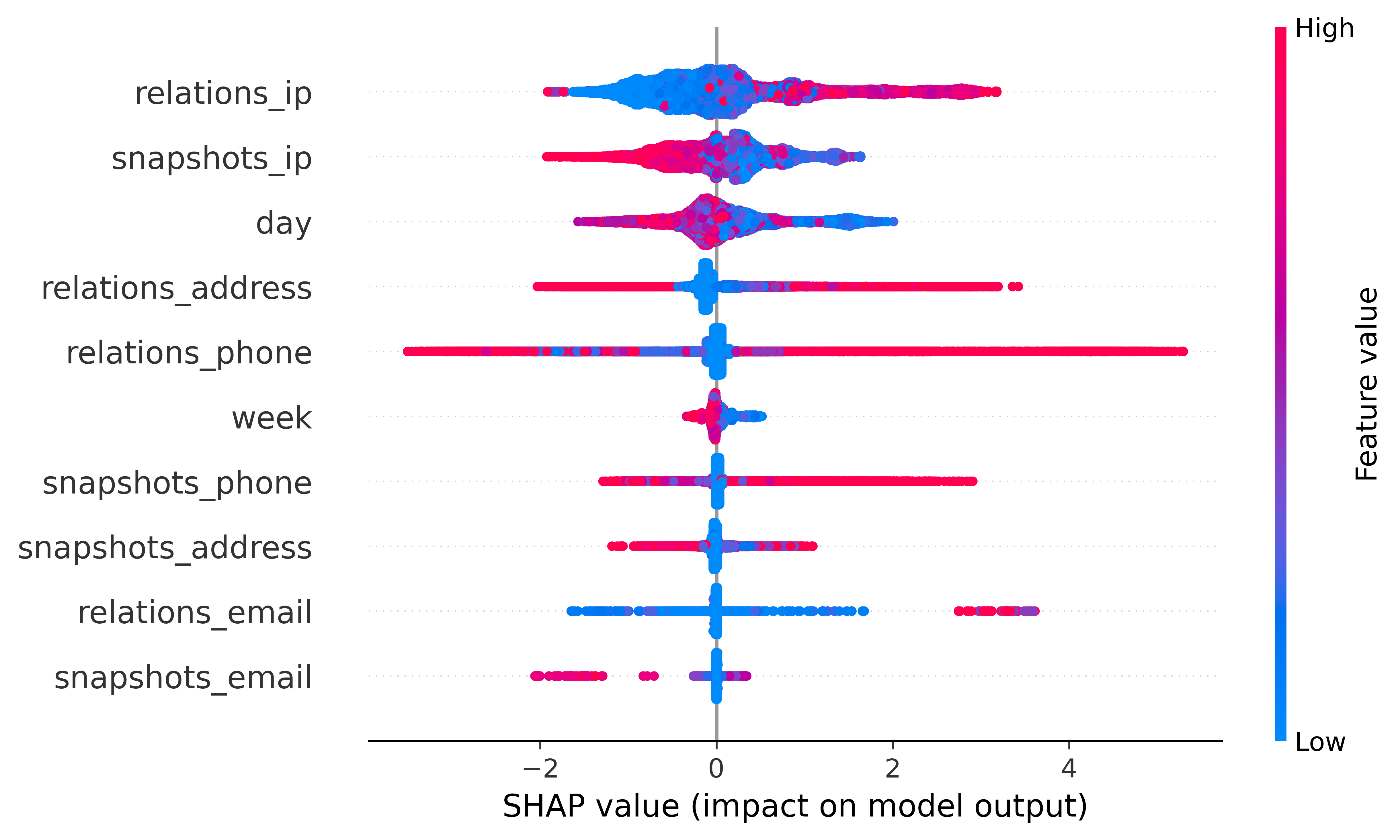}
    \caption{Shapley Values in XGBoost with Global Features.}
    \label{fig:shap}
\end{figure}

To assess the performance of our models, we decide to establish yet another series of ML baselines (Logistic regression, Random forest, XGBoost) using purely graph-derived features. Indeed, after the study of multiple graph features (e.g., Table~\ref{tab:features}) for the MassReg dataset, we observe for instance a correlation between the total number of relations of the linking entities (shipping address, phone number, ...) and the account riskiness (See Figure~\ref{fig:shap}). A natural explanation is that suspicious accounts are oftentimes registered using a common shipping address, such as a warehouse, or telephone numbers that are listed as spam calls. As a consequence, these linking entities will be linked to many accounts and are indicative of suspicious activities. 
Since in MassReg, DyHGN-* models perform the best, we only conduct the ML experiments on this dataset.

\textbf{Feature description.}
For each account in MassReg, we derive 8 features from the graph structural and temporal information. We provide an example to illustrate them in Table \ref{tab:features}. The row 1 in bold can be read as follows: the account n°1 was created on the $63^{th}$ day (in the $9^{th}$ week); it gave a shipping address and a phone number that were used respectively 4 and 9 different times, both appearing in two different time snapshots; this account also used an IP address that has been used twice and appeared in two different time snapshots; it did not provide any email address.

\textbf{Incremental setting.} 
In Table~\ref{tab:graph_features} we report the best ML performer XGBoost under various settings.
The global model in the Table \ref{tab:graph_features} relies on the features of the whole graph in all time stamps, which means that we take the features of the linking entities regardless of which time snapshot the account was created in. In contrast, we can also construct our same features incrementally, meaning that we only take the features from the graph that are available at the time of the account's creation.
As expected, the incremental setting performs worse than the global one. However, this is very common in graph-related tasks to investigate both cases (c.f. \cite{hu2020open}).

In the DyHGN-* architecture, we allow the edges and nodes in the test set to appear already in the training set, since the whole structural and temporal subgraphs are given as input to the GCN layers. This is a global setting. In the case of flagging frauds in a dynamic setting, both global and incremental settings are useful: the former allows us to pre-train a large model on historical+current data and capture fraudulent patterns with long-term dependencies, while the latter allows us to capture newly established fraudulent patterns better.

\textbf{Train/test split on the performance.}
In Table~\ref{tab:graph_features}, we report the results of our best ML performer XGBoost using two different types of train-test split. We observe a large discrepancy between them. For comparison, we also evaluated DyHGN on a random train/test split and found an AP score of about 0.89, which is about 7\% higher than the score with chronological split. While this confirms that a random split is overly optimistic compared to a chronological split, the discrepancy in the score observed with global and incremental features also indicates that it is actually hard to predict the future based on the past account creation logs. This aligns with our finding from Figure~\ref{fig:risky} in Appendix \ref{app:dist} that in MassReg, the uneven distribution of the labels across time makes it difficult to predict the future labels based on the past.

\textbf{Feature importance.}
One advantage of using handcrafted features is that this model can be better explained (vs. GNN explainability in \cite{rao2021xfraud}) and we can look at the Shapley (SHAP) values to closely identify the key features of the graph. 
The SHAP values shows the impact of each feature (y-axis ordered in descending order of feature importance), whether a feature contributes positively or negatively, to the prediction (x-axis). 
In Figure \ref{fig:shap}, we plot the SHAP values of the best performer (XGBoost with global features). Note that its performance is on par with the best DyHGN model (DyHGN-DE). The use frequency of one entity (e.g., IP, address, phone) is a crucial indicator of frauds. Also, the time dimension (e.g., day, week) is influential when flagging frauds. We see that if an account provides an IP address that has been used multiple times (high feature value of relations\_ip), it is more likely to be a risky account (impact on the model output leans toward the positive class). However, if the IP address appears in multiple time snapshots (snapshots\_ip), the account is less likely to be risky. Indeed, since we are concerned with massive registration of suspicious accounts here, we are mostly looking for IP addresses that would be used frequently in a restricted time period. The impact of snapshots\_email and relations\_email follows a similar pattern as IP. For shipping addresses and phone numbers, the interpretation is not as easy. We see that less used addresses and phone numbers are mostly linked with non-risky accounts, but they can be indicators of legit/fraud labels when they are highly used. On the temporal aspects, accounts created at the beginning (low day and week number) are more likely to be suspicious, which correlates with the label distribution of the MassReg dataset shown in Figure \ref{fig:risky}.
This resonates in the AP performances of DyHGN-DE, where long-term dependencies from the past can be learned.

To summarize, looking at ML models trained by graph-derived features helps us gain more insights of learning graph dynamics. It can guide us to model interpretability when using complex models such as HGNNs and DyHGN.  

\subsection{Discussions}
\label{sec:discussion}
We share thoughts about modelling graph dynamics with GCN, GAT, HGT and diachronic embeddings and discuss the architecture efficiency in prototyping and production.

\textbf{Modeling dynamic graphs with ``Attention".} \citet{lv2021we} discusses the efficiency and performances of 10 models (incl. GAT) on 7 datasets, but none of the models and datasets include graph dynamics. We are able to resonate to their findings by looking at graph dynamics. In line with the findings in ~\cite{lv2021we}, in datasets (xFraudTxn, xFraudAccount) with evenly distributed labels across time, GAT significantly outperforms models with HGT. Since the latter takes longer to train, it is beneficial to use GAT in such datasets with dynamics. For the dataset with unevenly distributed labels across time (MassReg), we have a better performance in our models (DyHGN-DE and DyHGN) over GAT.

\textbf{Efficiency.} Note that the largest DyHGN-* model can take between 4 and 15 min per epoch (on average 5.5 for MassReg, 14 for xFraudTxn, 9 for xFraudAccount), because we need to group by source nodes and the current implementation does not include batch training. But in the future, we plan to extend the implementation with (mini-)batch training once the prototype needs to run on an industrial-scale dataset as in xFraud \cite{rao2021xfraud}, where a deployment in a distributed setting is also explored and implemented.\\


\section{Related Work}\label{sec:lit}
We review several key areas relevant to our work.

\textbf{Heterogeneous Graph Neural Networks.}
Learning in heterogeneous graphs has gained interest in recent years, and recent works aim to generalize the traditional GCN and GAT to heterogeneous graphs~\cite{hu2020heterogeneous, wang2019han, hong2020attention, zhang2019hetgnn}. These models specify node types when constructing graphs and perform sampling over different types of nodes during message passing, which brings improvement in modelling heterogeneous graphs.  

\textbf{Dynamic Graph Neural Networks.}
The temporal dynamics are oftentimes investigated within a homogeneous graph setting. One representative work is DySAT~\cite{sankar2020dysat} that discusses an approach to learn deep neural representations on dynamic homogeneous graphs via self-attention networks. DySAT is applicable in cases where all entities in a graph have a dynamic perspective. However, in our application case, we need to differentiate between two types of entities in the time dimension: (1) accounts and transactions being dynamically added/removed across time, (2) hard linking entities like registration addresses and telephone numbers that stay static across time. 
Dynamic graphs could also be represented as a sequence of time events, which is the approach used in Temporal Graph Networks (TGN) \cite{rossi2020temporal}. Applying memory modules and graph-based operators, the model is able to outperform other approaches for link prediction in both transductive and inductive settings while being more computationally efficient.

\textbf{Fraud Detection.}
In the context of fraud detection, we already described DyHGN \cite{rao2020suspicious} that tackles suspicious massive registration detection with a homogeneous graph neural network that is composed of two subgraphs, a structural and a temporal and that can be applied to heterogeneous graphs.
xFraud \cite{rao2021xfraud} uses self-attentive heterogeneous graph neural network as predictor in a static setting and provides an explainer that generates meaningful insights to facilitate further process in business units.
Lambda Neural Network (LNN) \cite{lu2021graph} uses directed dynamic snapshot linkage design for graph construction, to ensure that information flow passed through neighbours only comes from the past. While LNN performs node classification in an inductive setting,
Asynchronous Propagation Attention Network (APAN) \cite{wang2021apan} adopts temporal encoding similar to TGN \cite{rossi2020temporal} and decouples graph computation and inference, to perform edge classification regarding fraudulent interactions.

\section{Conclusion}
In this paper, we investigate the benefits of building a dynamic heterogeneous graph neural network (DyHGN) with a diachronic entity embedding to provide characteristics of entities at any point in time. We also make the model heterogeneous and compare it with its homogeneous counterpart. We conduct experiments on three real-world graphs from application scenarios at eBay and benchmark several models on it. We find that our models are best-performers only in specific settings with uneven label distributions across time, and we also discuss a trade-off between performance and computation cost. In addition, we share insights on feature importance using Shapley values when investigating temporal graphs trained by XGBoost using graph-derived features. This interesting angle provides inspirations for future work in this direction.

\bibliographystyle{ACM-Reference-Format}
\bibliography{sample-base}


\begin{thebibliography}{24}


\ifx \showCODEN    \undefined \def \showCODEN     #1{\unskip}     \fi
\ifx \showDOI      \undefined \def \showDOI       #1{#1}\fi
\ifx \showISBNx    \undefined \def \showISBNx     #1{\unskip}     \fi
\ifx \showISBNxiii \undefined \def \showISBNxiii  #1{\unskip}     \fi
\ifx \showISSN     \undefined \def \showISSN      #1{\unskip}     \fi
\ifx \showLCCN     \undefined \def \showLCCN      #1{\unskip}     \fi
\ifx \shownote     \undefined \def \shownote      #1{#1}          \fi
\ifx \showarticletitle \undefined \def \showarticletitle #1{#1}   \fi
\ifx \showURL      \undefined \def \showURL       {\relax}        \fi
\providecommand\bibfield[2]{#2}
\providecommand\bibinfo[2]{#2}
\providecommand\natexlab[1]{#1}
\providecommand\showeprint[2][]{arXiv:#2}

\bibitem[Goel et~al\mbox{.}(2020)]%
        {goel2020diachronic}
\bibfield{author}{\bibinfo{person}{Rishab Goel}, \bibinfo{person}{Seyed~Mehran
  Kazemi}, \bibinfo{person}{Marcus Brubaker}, {and} \bibinfo{person}{Pascal
  Poupart}.} \bibinfo{year}{2020}\natexlab{}.
\newblock \showarticletitle{Diachronic embedding for temporal knowledge graph
  completion}. In \bibinfo{booktitle}{\emph{Proceedings of the AAAI Conference
  on Artificial Intelligence}}, Vol.~\bibinfo{volume}{34}.
  \bibinfo{pages}{3988--3995}.
\newblock


\bibitem[Hong et~al\mbox{.}(2020)]%
        {hong2020attention}
\bibfield{author}{\bibinfo{person}{Huiting Hong}, \bibinfo{person}{Hantao Guo},
  \bibinfo{person}{Yucheng Lin}, \bibinfo{person}{Xiaoqing Yang},
  \bibinfo{person}{Zang Li}, {and} \bibinfo{person}{Jieping Ye}.}
  \bibinfo{year}{2020}\natexlab{}.
\newblock \showarticletitle{An Attention-Based Graph Neural Network for
  Heterogeneous Structural Learning.}. In \bibinfo{booktitle}{\emph{AAAI}}.
  \bibinfo{pages}{4132--4139}.
\newblock


\bibitem[Hu et~al\mbox{.}(2020b)]%
        {hu2020open}
\bibfield{author}{\bibinfo{person}{Weihua Hu}, \bibinfo{person}{Matthias Fey},
  \bibinfo{person}{Marinka Zitnik}, \bibinfo{person}{Yuxiao Dong},
  \bibinfo{person}{Hongyu Ren}, \bibinfo{person}{Bowen Liu},
  \bibinfo{person}{Michele Catasta}, {and} \bibinfo{person}{Jure Leskovec}.}
  \bibinfo{year}{2020}\natexlab{b}.
\newblock \showarticletitle{Open graph benchmark: Datasets for machine learning
  on graphs}.
\newblock \bibinfo{journal}{\emph{Advances in neural information processing
  systems}}  \bibinfo{volume}{33} (\bibinfo{year}{2020}),
  \bibinfo{pages}{22118--22133}.
\newblock


\bibitem[Hu et~al\mbox{.}(2020a)]%
        {hu2020heterogeneous}
\bibfield{author}{\bibinfo{person}{Ziniu Hu}, \bibinfo{person}{Yuxiao Dong},
  \bibinfo{person}{Kuansan Wang}, {and} \bibinfo{person}{Yizhou Sun}.}
  \bibinfo{year}{2020}\natexlab{a}.
\newblock \showarticletitle{Heterogeneous graph transformer}. In
  \bibinfo{booktitle}{\emph{Proceedings of The Web Conference 2020}}.
  \bibinfo{pages}{2704--2710}.
\newblock


\bibitem[Li et~al\mbox{.}(2019)]%
        {li2019spam}
\bibfield{author}{\bibinfo{person}{Ao Li}, \bibinfo{person}{Zhou Qin},
  \bibinfo{person}{Runshi Liu}, \bibinfo{person}{Yiqun Yang}, {and}
  \bibinfo{person}{Dong Li}.} \bibinfo{year}{2019}\natexlab{}.
\newblock \showarticletitle{Spam review detection with graph convolutional
  networks}. In \bibinfo{booktitle}{\emph{Proceedings of the 28th ACM
  International Conference on Information and Knowledge Management}}.
  \bibinfo{pages}{2703--2711}.
\newblock


\bibitem[Liang et~al\mbox{.}(2019)]%
        {liang2019uncovering}
\bibfield{author}{\bibinfo{person}{Chen Liang}, \bibinfo{person}{Ziqi Liu},
  \bibinfo{person}{Bin Liu}, \bibinfo{person}{Jun Zhou},
  \bibinfo{person}{Xiaolong Li}, \bibinfo{person}{Shuang Yang}, {and}
  \bibinfo{person}{Yuan Qi}.} \bibinfo{year}{2019}\natexlab{}.
\newblock \showarticletitle{Uncovering Insurance Fraud Conspiracy with Network
  Learning}. In \bibinfo{booktitle}{\emph{Proceedings of the 42nd International
  ACM SIGIR Conference on Research and Development in Information Retrieval}}.
  \bibinfo{pages}{1181--1184}.
\newblock


\bibitem[Liu et~al\mbox{.}(2019)]%
        {liu2019geniepath}
\bibfield{author}{\bibinfo{person}{Ziqi Liu}, \bibinfo{person}{Chaochao Chen},
  \bibinfo{person}{Longfei Li}, \bibinfo{person}{Jun Zhou},
  \bibinfo{person}{Xiaolong Li}, \bibinfo{person}{Le Song}, {and}
  \bibinfo{person}{Yuan Qi}.} \bibinfo{year}{2019}\natexlab{}.
\newblock \showarticletitle{Geniepath: Graph neural networks with adaptive
  receptive paths}. In \bibinfo{booktitle}{\emph{Proceedings of the AAAI
  Conference on Artificial Intelligence}}, Vol.~\bibinfo{volume}{33}.
  \bibinfo{pages}{4424--4431}.
\newblock


\bibitem[Liu et~al\mbox{.}(2018)]%
        {liu2018heterogeneous}
\bibfield{author}{\bibinfo{person}{Ziqi Liu}, \bibinfo{person}{Chaochao Chen},
  \bibinfo{person}{Xinxing Yang}, \bibinfo{person}{Jun Zhou},
  \bibinfo{person}{Xiaolong Li}, {and} \bibinfo{person}{Le Song}.}
  \bibinfo{year}{2018}\natexlab{}.
\newblock \showarticletitle{Heterogeneous graph neural networks for malicious
  account detection}. In \bibinfo{booktitle}{\emph{Proceedings of the 27th ACM
  International Conference on Information and Knowledge Management}}.
  \bibinfo{pages}{2077--2085}.
\newblock


\bibitem[Lu et~al\mbox{.}(2021)]%
        {lu2021graph}
\bibfield{author}{\bibinfo{person}{Mingxuan Lu}, \bibinfo{person}{Zhichao Han},
  \bibinfo{person}{Zitao Zhang}, \bibinfo{person}{Yang Zhao}, {and}
  \bibinfo{person}{Yinan Shan}.} \bibinfo{year}{2021}\natexlab{}.
\newblock \showarticletitle{Graph Neural Networks in Real-Time Fraud Detection
  with Lambda Architecture}.
\newblock \bibinfo{journal}{\emph{arXiv preprint arXiv:2110.04559}}
  (\bibinfo{year}{2021}).
\newblock


\bibitem[Lv et~al\mbox{.}(2021)]%
        {lv2021we}
\bibfield{author}{\bibinfo{person}{Qingsong Lv}, \bibinfo{person}{Ming Ding},
  \bibinfo{person}{Qiang Liu}, \bibinfo{person}{Yuxiang Chen},
  \bibinfo{person}{Wenzheng Feng}, \bibinfo{person}{Siming He},
  \bibinfo{person}{Chang Zhou}, \bibinfo{person}{Jianguo Jiang},
  \bibinfo{person}{Yuxiao Dong}, {and} \bibinfo{person}{Jie Tang}.}
  \bibinfo{year}{2021}\natexlab{}.
\newblock \showarticletitle{Are we really making much progress? Revisiting,
  benchmarking, and refining heterogeneous graph neural networks}.
\newblock  (\bibinfo{year}{2021}).
\newblock


\bibitem[Ma et~al\mbox{.}(2018)]%
        {ma2018graphrad}
\bibfield{author}{\bibinfo{person}{Jun Ma}, \bibinfo{person}{Danqing Zhang},
  \bibinfo{person}{Yun Wang}, \bibinfo{person}{Yan Zhang}, {and}
  \bibinfo{person}{Alexey Pozdnoukhov}.} \bibinfo{year}{2018}\natexlab{}.
\newblock \showarticletitle{GraphRAD: A Graph-based Risky Account Detection
  System}.
\newblock  (\bibinfo{year}{2018}).
\newblock


\bibitem[Rao et~al\mbox{.}(2022)]%
        {rao2021xfraud}
\bibfield{author}{\bibinfo{person}{Susie~Xi Rao}, \bibinfo{person}{Shuai
  Zhang}, \bibinfo{person}{Zhichao Han}, \bibinfo{person}{Zitao Zhang},
  \bibinfo{person}{Wei Min}, \bibinfo{person}{Zhiyao Chen},
  \bibinfo{person}{Yinan Shan}, \bibinfo{person}{Yang Zhao}, {and}
  \bibinfo{person}{Ce Zhang}.} \bibinfo{year}{2022}\natexlab{}.
\newblock \showarticletitle{xFraud: Explainable Fraud Transaction Detection}.
\newblock \bibinfo{journal}{\emph{VLDB}} (\bibinfo{year}{2022}).
\newblock
Issue 3.
\urldef\tempurl%
\url{https://doi.org/10.14778/3494124.3494128}
\showDOI{\tempurl}


\bibitem[Rao et~al\mbox{.}(2020)]%
        {rao2020suspicious}
\bibfield{author}{\bibinfo{person}{Susie~Xi Rao}, \bibinfo{person}{Shuai
  Zhang}, \bibinfo{person}{Zhichao Han}, \bibinfo{person}{Zitao Zhang},
  \bibinfo{person}{Wei Min}, \bibinfo{person}{Mo Cheng}, \bibinfo{person}{Yinan
  Shan}, \bibinfo{person}{Yang Zhao}, {and} \bibinfo{person}{Ce Zhang}.}
  \bibinfo{year}{2020}\natexlab{}.
\newblock \showarticletitle{Suspicious Massive Registration Detection via
  Dynamic Heterogeneous Graph Neural Networks}.
\newblock \bibinfo{journal}{\emph{arXiv preprint arXiv:2012.10831}}
  (\bibinfo{year}{2020}).
\newblock


\bibitem[Rossi et~al\mbox{.}(2020)]%
        {rossi2020temporal}
\bibfield{author}{\bibinfo{person}{Emanuele Rossi}, \bibinfo{person}{Ben
  Chamberlain}, \bibinfo{person}{Fabrizio Frasca}, \bibinfo{person}{Davide
  Eynard}, \bibinfo{person}{Federico Monti}, {and} \bibinfo{person}{Michael
  Bronstein}.} \bibinfo{year}{2020}\natexlab{}.
\newblock \showarticletitle{Temporal graph networks for deep learning on
  dynamic graphs}.
\newblock \bibinfo{journal}{\emph{arXiv preprint arXiv:2006.10637}}
  (\bibinfo{year}{2020}).
\newblock


\bibitem[Sankar et~al\mbox{.}(2020)]%
        {sankar2020dysat}
\bibfield{author}{\bibinfo{person}{Aravind Sankar}, \bibinfo{person}{Yanhong
  Wu}, \bibinfo{person}{Liang Gou}, \bibinfo{person}{Wei Zhang}, {and}
  \bibinfo{person}{Hao Yang}.} \bibinfo{year}{2020}\natexlab{}.
\newblock \showarticletitle{Dysat: Deep neural representation learning on
  dynamic graphs via self-attention networks}. In
  \bibinfo{booktitle}{\emph{Proceedings of the 13th International Conference on
  Web Search and Data Mining}}. \bibinfo{pages}{519--527}.
\newblock


\bibitem[Shekhar et~al\mbox{.}(2020)]%
        {shekhar2020entity}
\bibfield{author}{\bibinfo{person}{Shubhranshu Shekhar},
  \bibinfo{person}{Deepak Pai}, {and} \bibinfo{person}{Sriram Ravindran}.}
  \bibinfo{year}{2020}\natexlab{}.
\newblock \showarticletitle{Entity Resolution in Dynamic Heterogeneous
  Networks}. In \bibinfo{booktitle}{\emph{Companion Proceedings of the Web
  Conference 2020}}. \bibinfo{pages}{662--668}.
\newblock


\bibitem[Wang et~al\mbox{.}(2019b)]%
        {wang2019fdgars}
\bibfield{author}{\bibinfo{person}{Jianyu Wang}, \bibinfo{person}{Rui Wen},
  \bibinfo{person}{Chunming Wu}, \bibinfo{person}{Yu Huang}, {and}
  \bibinfo{person}{Jian Xion}.} \bibinfo{year}{2019}\natexlab{b}.
\newblock \showarticletitle{Fdgars: Fraudster detection via graph convolutional
  networks in online app review system}. In \bibinfo{booktitle}{\emph{Companion
  Proceedings of The 2019 World Wide Web Conference}}.
  \bibinfo{pages}{310--316}.
\newblock


\bibitem[Wang et~al\mbox{.}(2019a)]%
        {wang2019han}
\bibfield{author}{\bibinfo{person}{Xiao Wang}, \bibinfo{person}{Houye Ji},
  \bibinfo{person}{Chuan Shi}, \bibinfo{person}{Bai Wang},
  \bibinfo{person}{Yanfang Ye}, \bibinfo{person}{Peng Cui}, {and}
  \bibinfo{person}{Philip~S Yu}.} \bibinfo{year}{2019}\natexlab{a}.
\newblock \showarticletitle{Heterogeneous graph attention network}. In
  \bibinfo{booktitle}{\emph{The World Wide Web Conference}}.
  \bibinfo{pages}{2022--2032}.
\newblock


\bibitem[Wang et~al\mbox{.}(2021)]%
        {wang2021apan}
\bibfield{author}{\bibinfo{person}{Xuhong Wang}, \bibinfo{person}{Ding Lyu},
  \bibinfo{person}{Mengjian Li}, \bibinfo{person}{Yang Xia},
  \bibinfo{person}{Qi Yang}, \bibinfo{person}{Xinwen Wang},
  \bibinfo{person}{Xinguang Wang}, \bibinfo{person}{Ping Cui},
  \bibinfo{person}{Yupu Yang}, \bibinfo{person}{Bowen Sun}, {et~al\mbox{.}}}
  \bibinfo{year}{2021}\natexlab{}.
\newblock \showarticletitle{APAN: Asynchronous Propagation Attention Network
  for Real-time Temporal Graph Embedding}. In
  \bibinfo{booktitle}{\emph{Proceedings of the 2021 International Conference on
  Management of Data}}. \bibinfo{pages}{2628--2638}.
\newblock


\bibitem[Weber et~al\mbox{.}(2019)]%
        {weber2019anti}
\bibfield{author}{\bibinfo{person}{Mark Weber}, \bibinfo{person}{Giacomo
  Domeniconi}, \bibinfo{person}{Jie Chen}, \bibinfo{person}{Daniel Karl~I
  Weidele}, \bibinfo{person}{Claudio Bellei}, \bibinfo{person}{Tom Robinson},
  {and} \bibinfo{person}{Charles~E Leiserson}.}
  \bibinfo{year}{2019}\natexlab{}.
\newblock \showarticletitle{Anti-money laundering in bitcoin: Experimenting
  with graph convolutional networks for financial forensics}.
\newblock \bibinfo{journal}{\emph{arXiv preprint arXiv:1908.02591}}
  (\bibinfo{year}{2019}).
\newblock


\bibitem[Wen et~al\mbox{.}(2020)]%
        {wen2020asa}
\bibfield{author}{\bibinfo{person}{Rui Wen}, \bibinfo{person}{Jianyu Wang},
  \bibinfo{person}{Chunming Wu}, {and} \bibinfo{person}{Jian Xiong}.}
  \bibinfo{year}{2020}\natexlab{}.
\newblock \showarticletitle{ASA: Adversary Situation Awareness via
  Heterogeneous Graph Convolutional Networks}. In
  \bibinfo{booktitle}{\emph{Companion Proceedings of the Web Conference 2020}}.
  \bibinfo{pages}{674--678}.
\newblock


\bibitem[Yang et~al\mbox{.}(2014)]%
        {yang2014embedding}
\bibfield{author}{\bibinfo{person}{Bishan Yang}, \bibinfo{person}{Wen-tau Yih},
  \bibinfo{person}{Xiaodong He}, \bibinfo{person}{Jianfeng Gao}, {and}
  \bibinfo{person}{Li Deng}.} \bibinfo{year}{2014}\natexlab{}.
\newblock \showarticletitle{Embedding entities and relations for learning and
  inference in knowledge bases}.
\newblock \bibinfo{journal}{\emph{arXiv preprint arXiv:1412.6575}}
  (\bibinfo{year}{2014}).
\newblock


\bibitem[Zhang et~al\mbox{.}(2019b)]%
        {zhang2019hetgnn}
\bibfield{author}{\bibinfo{person}{Chuxu Zhang}, \bibinfo{person}{Dongjin
  Song}, \bibinfo{person}{Chao Huang}, \bibinfo{person}{Ananthram Swami}, {and}
  \bibinfo{person}{Nitesh~V Chawla}.} \bibinfo{year}{2019}\natexlab{b}.
\newblock \showarticletitle{Heterogeneous graph neural network}. In
  \bibinfo{booktitle}{\emph{Proceedings of the 25th ACM SIGKDD International
  Conference on Knowledge Discovery \& Data Mining}}.
  \bibinfo{pages}{793--803}.
\newblock


\bibitem[Zhang et~al\mbox{.}(2019a)]%
        {zhang2019key}
\bibfield{author}{\bibinfo{person}{Yiming Zhang}, \bibinfo{person}{Yujie Fan},
  \bibinfo{person}{Yanfang Ye}, \bibinfo{person}{Liang Zhao}, {and}
  \bibinfo{person}{Chuan Shi}.} \bibinfo{year}{2019}\natexlab{a}.
\newblock \showarticletitle{Key Player Identification in Underground Forums
  over Attributed Heterogeneous Information Network Embedding Framework}. In
  \bibinfo{booktitle}{\emph{Proceedings of the 28th ACM International
  Conference on Information and Knowledge Management}}.
  \bibinfo{pages}{549--558}.
\newblock


\end{thebibliography}

\clearpage
\appendix

\section{Experiment Results using AUC score}\label{app:auc}

We report the AUC scores of our models and their baselines in Table \ref{tab:result_AUC}. The AUC score corresponds to the Area Under the Curve of the Receiver Operating Characteristic, which is obtained by plotting the true positive rate against the false positive rate at various thresholds. We make the following observations:
\begin{enumerate}[leftmargin=\parindent]
    \item GAT outperforms DyHGN-* models according to the AUC metric for the MassReg dataset, while DyHGN-* models show better performances with the AP score in Table \ref{tab:result}. We observe that the differences in AUC scores between  models on all datasets are tighter than those in the AP metric. However, we prefer the AP score, since it is more sensitive to the improvements for the positive class, which is our main focus. For xFraudTxn and xFraudAccount, GAT and Simple-HGN remain the best performer, as the AP scores indicate. 
    \item Compared with the results in \citet{rao2021xfraud}, our AUC is higher because we run more epochs on the models. However, this does come with a cost of having more complicated design of graph construction and model architecture. Therefore, when one wants to model the graph dynamics, one should do it with ``attention" (!) and benchmark different solutions against the computation cost.
\end{enumerate}

\begin{table}[!h]
\centering
\resizebox{\linewidth}{!}{\begin{tabular}{cccc}
\toprule
&MassReg & xFraudTxn & xFraudAccount \\
\midrule
GCN      & 0.8627 $\pm$ 0.0044 & 0.7647 $\pm$ 0.0078 & 0.5793 $\pm$ 0.0284 \\
GAT      & \textbf{0.8806} $\pm$ 0.0026 & \textbf{0.8880} $\pm$ 0.0066 & 0.7572 $\pm$ 0.0178 \\
Simple-HGN  & 0.8260 $\pm$ 0.0204 & 0.6124 $\pm$ 0.0316 & \textbf{0.8456} $\pm$ 0.0270 \\
DyHGN   & 0.8749 $\pm$ 0.0059 & 0.7878 $\pm$ 0.0060 & 0.7527 $\pm$ 0.0733 \\
DyHGN-DE  & 0.8793 $\pm$ 0.0031 & 0.6288 $\pm$ 0.0065 & 0.6779 $\pm$ 0.0081 \\
DyHGN-DE-HGT  & 0.8600 $\pm$ 0.0078 & 0.6263 $\pm$ 0.0175 & 0.6851 $\pm$ 0.0051 \\
\bottomrule
\end{tabular}}
\caption{AUC Scores. ``DE": diachronic embedding, ``HGT": heterogeneous graph transformer.}
\label{tab:result_AUC}
\end{table}

\section{Hyperparameters}\label{app:hyper}
\begin{table*}[!t]
\centering
\resizebox{\linewidth}{!}{\begin{tabular}{cccccccccccccccc}
\toprule
\textbf{Model} & \multicolumn{3}{c}{\textbf{n\_layer}}                           & \multicolumn{3}{c}{\textbf{n\_hid}}                             & \textbf{dropout} & \textbf{optimizer} & \textbf{lr} & \textbf{max\_epochs} & \textbf{patience} & \textbf{n\_heads} & \multicolumn{3}{c}{\textbf{diachronic\_embedding}}              \\ \midrule
               & \multicolumn{1}{l}{MassReg} & \multicolumn{1}{l}{xFraudTxn} & Account  & \multicolumn{1}{l}{MassReg} & \multicolumn{1}{l}{xFraudTxn} & Account &                  &                    &             &                      &                   &                   & \multicolumn{1}{l}{MassReg} & xFraudTxn & \multicolumn{1}{l}{xFraudAccount} \\ \cline{14-16} \cline{2-7}
               \\ 
GCN            & 4                           & 4                       & 4       & 256                         & 256                     & 256     & 0.1              & adamw              & 0.001       & 2048                 & 64                & -                 & -                           & -   & -                           \\
GAT            & 8                           & 2                       & 2       & 256                         & 256                     & 128     & 0.1              & adamw              & 0.001       & 2048                 & 64                & 4                 & -                           & -   & -                           \\
Simple-HGN     & 2                           & 2                       & 2       & 256                         & 64                      & 256     & 0.1              & admaw              & 0.001       & 2048                 & 64                & 4                 & -                           & -   & -                           \\
DyHGN         & 4                           & 2                       & 4       & 256                         & 256                     & 128     & 0.1              & adamw              & 0.001       & 2048                 & 64                & -                 & -                           & -   & -                           \\
DyHGN-DE      & 4                           & 2                       & 2       & 256                         & 128                     & 128     & 0.1              & adamw              & 0.001       & 128                  & 64                & -                 & 60                          & 10  & 10                          \\
DyHGN-DE-HGT  & 4                           & 2                       & 2       & 256                         & 128                     & 128     & 0.1              & adamw              & 0.001       & 128                  & 64                & 4                 & 30                          & 10  & 10                          \\ \bottomrule
\end{tabular}}
\caption{Hyperparameters on our dataset.  ``DE": diachronic embedding, ``HGT": heterogeneous graph transformer.}
\label{tab:hyperparam}
\end{table*}

We list the hyperparameters in the GNN models in Table~\ref{tab:hyperparam} and the XGBoost best parameters are colsample\_bytree: 0.9, eval\_metric: error, gamma: 0, learning\_rate: 0.1, max\_depth: 7,
min\_child\_weight: 5, n\_estimators: 200, objective: binary:logistic, reg\_alpha: 0, reg\_lam-
bda: 1, scale\_pos\_weight: 1, subsample: 0.7.

\section{Distribution of risky nodes across time}\label{app:dist}
\begin{figure*}[!t]
    \centering
    \begin{tabular}{ccc}
        \includegraphics[width=0.32\textwidth]{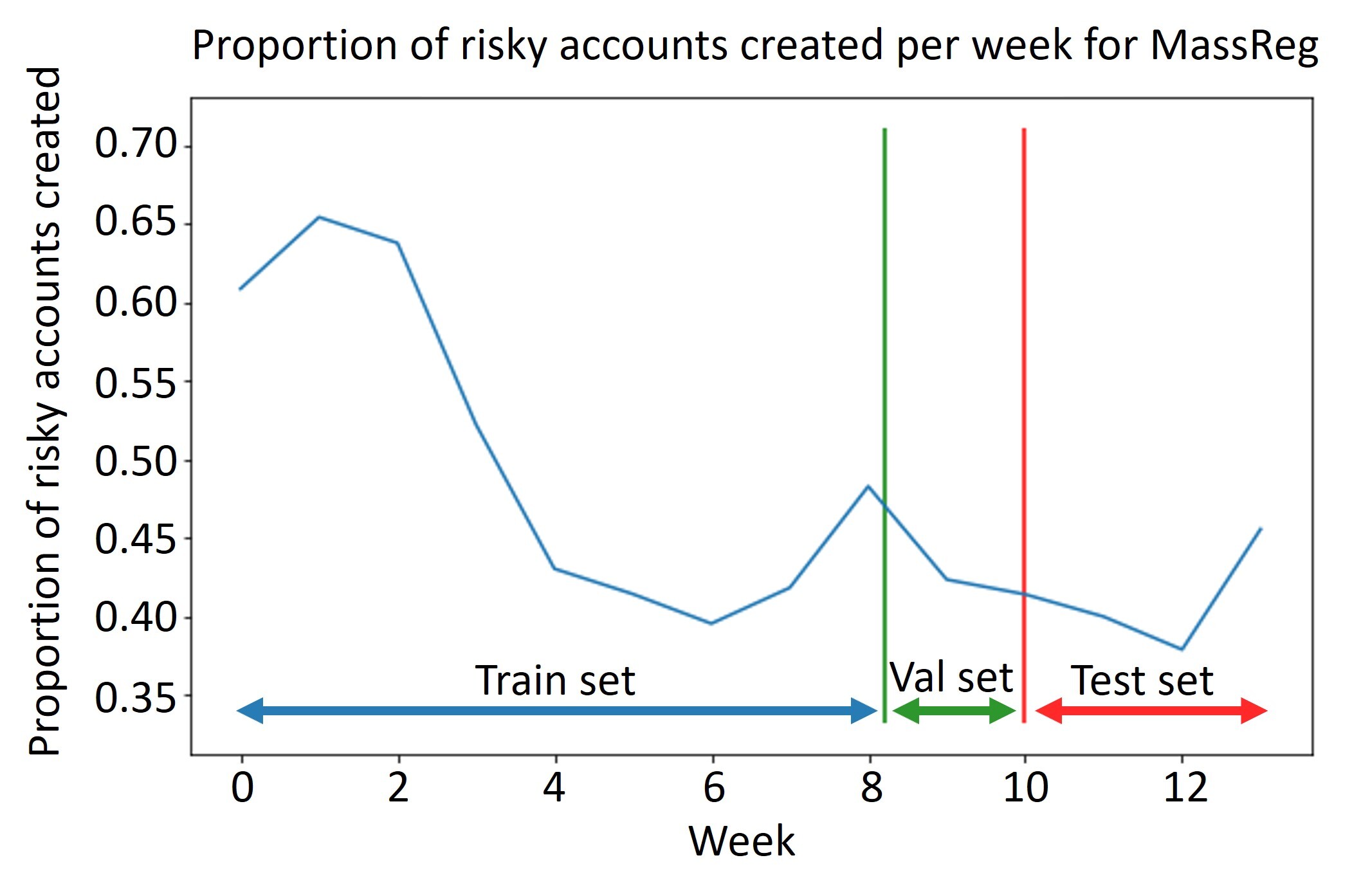} & 
        \includegraphics[width=0.33\textwidth]{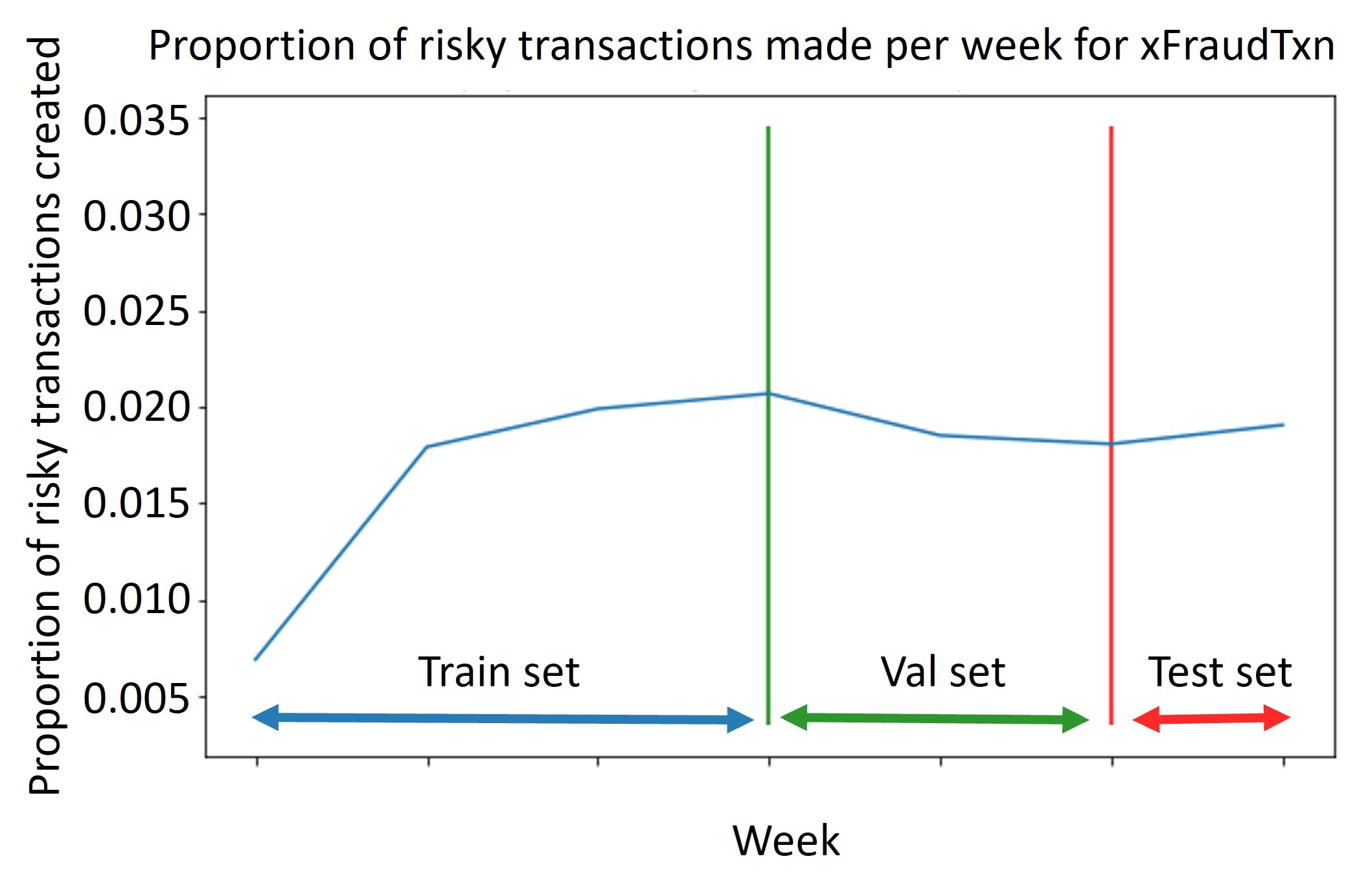} &
        \includegraphics[width=0.34\textwidth]{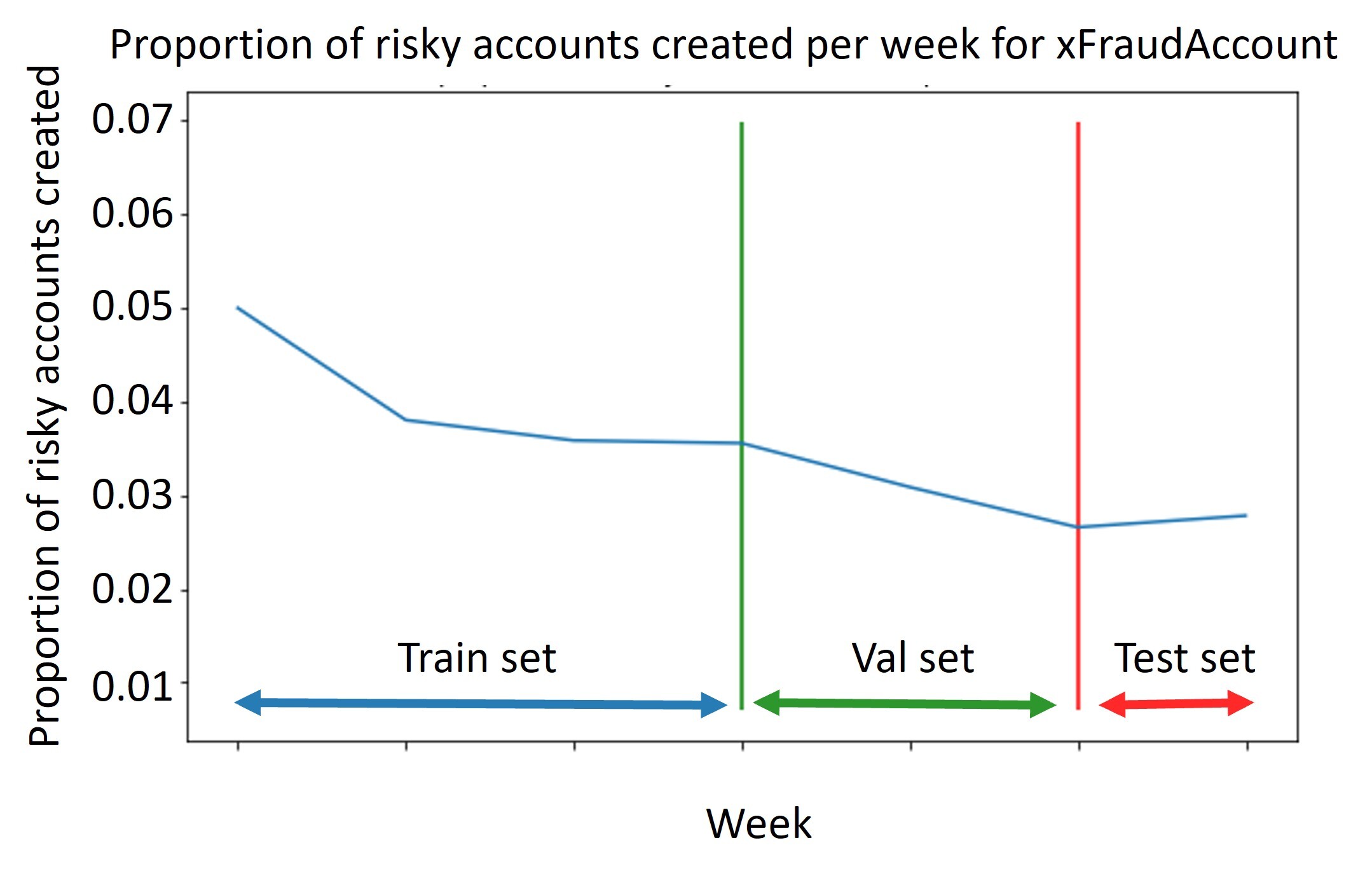}\\
    \end{tabular}
    \caption{Distribution of risky accounts across time for the MassReg dataset and the two xFraud datasets.}
    \label{fig:risky}
\end{figure*}

Figure \ref{fig:risky} shows the distribution of labels across time for the three datasets. We observe that the MassReg dataset is particularly unevenly distributed, the proportion of risky accounts varies from 40\% to 65\% and has two peaks at weeks 2 and 8 (the peak at week 13 can be discarded since this week was truncated and thus includes fewer accounts). On the contrary, the xFraud datasets are more evenly distributed. We believe this major difference to be the reason why certain models work better on the MassReg dataset but not on the xFraud datasets. Indeed, in the case of unevenly distributed datasets, the diachronic embeddings from the previous time stamps might be helpful in learning long-distance dependencies.


\begin{figure*}[!t]
    \centering
    \begin{tabular}{cc}
    \includegraphics[width=0.4\textwidth]{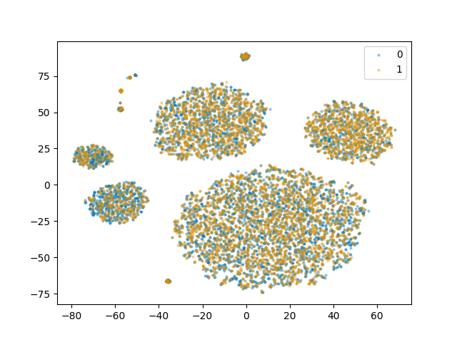} &
    \includegraphics[width=0.4\textwidth]{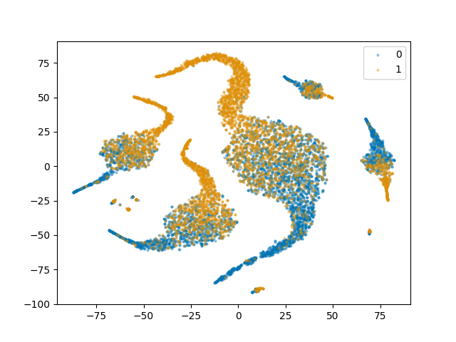} \\
    \textbf{Epoch 0 - AP 0.430} & \textbf{Epoch 170 - AP 0.815}
    \end{tabular}
    \caption{t-SNE visualization of the diachronic embedding at the beginning and at the end of the training.}
    \label{fig:embedding}
\end{figure*}

\section{Other Experiments concerning the Diachronic Embedding} \label{app:diachronic-emb}
In order to fully investigate the potential effect of the diachronic embedding, we carried supplementary ablation studies to the ones already described in Section \ref{sec:ablation}. It is important to notice that these experiments were only carried on the MassReg dataset, since DyHGN-* models have outperformed the GNN baselines in AP scores.

\textbf{Types of scores.}
In MassReg, when we consider a relation between an account (source node) and a linking entity (address, email, phone number, IP address) the information carried by the edge is redundant with the information carried by the linking entity (target node) since the edge already indicates the type of the target node. Therefore, we tried to compute the DE-DistMult score with the source node and the relation only. This gave us a boost in terms of computation time per epoch (about 3 minutes per epoch); however, we observed a decrease in performance. This could stem from the fact that target nodes are used multiple times and so the diachronic embedding could also use this information which is beneficial to the model.

\textbf{Embedding size.}
We experimented with different embedding sizes, and we found that a larger size of embedding seems to entail a higher score. However, increasing the embedding size is computationally heavy both in terms of space and of time, therefore, one needs to find a trade-off between complexity and performance.

\textbf{Visualization.}
We show some visualization of the evolution of the diachronic embedding throughout training in Figure \ref{fig:embedding}. The visualization is obtained by using t-SNE technique on the diachronic embedding at different epochs of the training. Each point represents an account of the MassReg dataset. The color of the point indicates the label of the account (0/blue for non-risky and 1/orange for risky). 

We observe that at epoch 0, there is no distinction between the two labels. The clusters that we observe are actually indicative of the number of linking entities (email, phone number, address, etc.) that the account has. These clusters are natural, since the diachronic embedding of a given account is obtained by averaging the scores of all the relations of this account.
Then, at epoch 170, we see that the two classes are partly separated, which indicates that the diachronic embedding was able to learn to classify the accounts to some extent. However, we also observe that many accounts are still mixed with accounts of the opposite label, which might explain why the diachronic embedding does not improve the performance of the model very significantly. We also see different degrees of separation across clusters (linking entities), this corresponds to the feature importance (SHAP values and their impacts on prediction) we discussed in Figure~\ref{fig:shap} in Sec.~\ref{sec:gnn-vs-ml}.

\end{document}